\documentclass[sigconf]{acmart}

\usepackage[capitalize,noabbrev]{cleveref}

\theoremstyle{plain}

\theoremstyle{definition}

\theoremstyle{remark}

\usepackage[textsize=tiny]{todonotes}

\usepackage[utf8]{inputenc} 
\usepackage{hyperref}       
\usepackage{url}            
\usepackage{booktabs}       
\usepackage{amsfonts}       
\usepackage{nicefrac}       
\usepackage{microtype}      
\usepackage{xcolor}         

\usepackage{graphicx}

\usepackage[utf8]{inputenc}

\usepackage{booktabs} %
\usepackage{amsfonts} %
\usepackage{nicefrac} %
\usepackage{microtype} %
\usepackage{graphicx}
\usepackage{natbib}
\usepackage{gensymb}
\usepackage{color}
\usepackage{caption}
\usepackage{subcaption}
\usepackage{amsthm}
\usepackage{mathrsfs}
\usepackage{amsmath}
\usepackage{mathtools}
\usepackage{wrapfig}
\usepackage{soul}
\usepackage{marvosym}
\usepackage[ruled,longend, algo2e]{algorithm2e}
\usepackage{algorithm}
\usepackage{algorithmic}
\usepackage{hhline}
\usepackage{multirow}

\usepackage{stackengine}

\makeatletter
\newcommand{\distas}[1]{\mathbin{\overset{#1}{\kern\z@\sim}}}%

\newcommand{\beqs}{\vspace{0mm}\begin{eqnarray}}
\newcommand{\eeqs}{\vspace{0mm}\end{eqnarray}}
\newcommand{\barr}{\begin{array}}
\newcommand{\earr}{\end{array}}

\def\rr{\textcolor{red}}

\usepackage{wrapfig}
\usepackage{multirow}

\usepackage[utf8]{inputenc} 
\usepackage[T1]{fontenc}    
\usepackage{hyperref}       
\usepackage{url}            
\usepackage{booktabs}       
\usepackage{amsfonts}       
\usepackage{nicefrac}       
\usepackage{microtype}      
\usepackage{xcolor}         

\usepackage[marginal]{footmisc}
\usepackage{makecell}
\renewcommand{\thefootnote}

\AtBeginDocument{%
  }

\setcopyright{acmlicensed}
\copyrightyear{2025}
\acmYear{2025}
\acmDOI{3746027.3755641}
\acmConference[MM '25] {Proceedings of the 33rd ACM International Conference on Multimedia}{October 27--31, 2025}{Dublin, Ireland.}
\acmBooktitle{Proceedings of the 33rd ACM International Conference on Multimedia (MM '25), October 27--31, 2025, Dublin, Ireland}
\acmISBN{979-8-4007-2035-2/2025/10}

\begin{document}

\title{SP-Mamba: Spatial-Perception State Space Model for Unsupervised Medical Anomaly Detection}


\author{Rui Pan}
\orcid{https://orcid.org/0009-0003-7767-4265}
\affiliation{%
  \institution{Faculty of Integrated Circuit, Xidian University}
  \city{Chang'an Qu}
  \state{Xi'an Shi}
  \country{China}}
\email{ruipan@stu.xidian.edu.cn}

\author{Ruiying Lu}
\authornote{Corresponding author}
\orcid{https://orcid.org/0000-0002-8825-6064}
\affiliation{%
  \institution{School of Cyber Engineering, Xidian University}
  \city{Chang'an Qu}
  \state{Xi'an Shi}
  \country{China}}
\email{luruiying@xidian.edu.cn}

\renewcommand{\shortauthors}{Rui Pan and Ruiying Lu}

\begin{abstract}
Radiography imaging protocols target on specific anatomical regions, resulting in highly consistent images with recurrent structural patterns across patients. Recent advances in medical anomaly detection have demonstrated the effectiveness of CNN- and transformer-based approaches. However, CNNs exhibit limitations in capturing long-range dependencies, while transformers suffer from quadratic computational complexity. In contrast, Mamba-based models, leveraging superior long-range modeling, structural feature extraction, and linear computational efficiency, have emerged as a promising alternative. To capitalize on the inherent structural regularity of medical images, this study introduces SP-Mamba, a spatial-perception Mamba framework for unsupervised medical anomaly detection. The window-sliding prototype learning and Circular-Hilbert scanning-based Mamba are introduced to better exploit consistent anatomical patterns and leverage spatial information for medical anomaly detection. Furthermore, we excavate the concentration and contrast characteristics of anomaly maps for improving anomaly detection. Extensive experiments on three diverse medical anomaly detection benchmarks confirm the proposed method's state-of-the-art performance, validating its efficacy and robustness. The code is available at \href{https://github.com/Ray-RuiPan/SP-Mamba}{https://github.com/Ray-RuiPan/SP-Mamba}.
\end{abstract}

\ccsdesc[100]{Computing methodologies}
\ccsdesc[500]{Computing methodologies~Scene anomaly detection}

\keywords{Medical anomaly detection, State space model, Mamba}


\maketitle

\section{Introduction}

Medical anomaly detection (AD), as a critical branch of medical artificial intelligence, is dedicated to identifying subtle deviations from normal patterns in healthcare data, preventing misdiagnoses and enabling timely early interventions~\cite{fernando2021deep,zhang2020viral}. The primary objective of visual medical AD is to identify images containing diseases and pinpoint anomalous pixels within them. However, a fundamental challenge stems from obtaining anomalous samples that comprehensively represent the high-dimensional disease types continuum, particularly given that the expert-dependent specialized annotations~\cite{cai2022dual} are necessary. This practical constraint has driven the paradigm shift toward formulating AD tasks as one-class classification problems, wherein only normal data is utilized for model training. Typically, prevailing methodologies predominantly adopt reconstruction-based frameworks, where anomaly scoring derives from per-pixel reconstruction error metrics or divergence from learned probability distributions. While recent advancements in unsupervised medical AD have demonstrated promising results, there still remain challenges for optimizing the accuracy-efficiency trade-off.

Detecting medical anomalies with high accuracy and efficiency is challenging. Currently, CNN-based methods and Transformer-based methods dominate this field of medical anomaly detection. While CNN-based methods excel at capturing local contextual features, they lack the ability to inherently model long-range dependencies. In contrast,  Transformers demonstrate superior global dependency modeling. Recent advancements, such as SQUID~\cite{xiang_squid_2023} and SimSID~\cite{xiang_exploiting_2024}, have achieved state-of-the-art (SOTA) performances in unsupervised medical anomaly detection by incorporating space-aware memory queues into Transformer architectures for inpainting and identifying anomalies in radiography images. Despite their superior performance and global modeling capabilities, Transformer-based methods suffer from quadratic computational complexity, constrained to process only low-resolution feature maps, which may adversely affect detection performance.

Recent advances in Mamba~\cite{gu2023mamba} show exceptional performance in language models, offering linear complexity versus Transformers' quadratic scaling while maintaining effectiveness. Mamba's introduction to vision tasks has triggered an explosion of research activity~\cite{liu2024vmamba,shi2025vmambair,ruan2024vm,wang2024mamba}. Recently, MambaAD pioneered the application of Mamba in the anomaly detection of photography images, leveraging linear complexity to compute anomaly maps. However, key differences exist between photography and radiography imaging. Photographic images assume translation invariance, where an object's meaning persists across various locations. Radiography imaging, conversely, relies on precise anatomical positioning, where structure and orientation critically indicate normal versus pathological states~\cite{haghighi2021transferable,xiang_squid_2023}.  It is critical and challenging to exploit consistent anatomical patterns and leverage spatial information into Mamba for anomaly detection for radiography images.

In this paper, we propose a spatial-perception state space model, named SP-Mamba, exquisitely designed for unsupervised medical anomaly detection. Specifically, SP-Mamba employs a pyramid-structured auto-encoder, where the Mamba-based decoder aims to reconstruct multi-scale normal features accurately, while leading to large reconstruction errors for anomalies. Specifically, in encoding space, we learn medical typical prototypes for radiography images and constraints within an adaptive window, to spatially capture consistent anatomical patterns in radiography images. During decoding, we introduce Circular-Hilbert scanning-based Mamba to better exploit spatial information for anomaly detection. Furthermore, we explore the concentration and contrast characteristics of anomaly maps in medical images and leverage them for better anomaly detection. Our contributions can be summarized as follows:
\begin{itemize}
    \item We propose an innovative mamba-based model to address unsupervised medical anomaly detection tasks, where the Circular-Hilbert scanning mechanism is proposed to fully capture the spatial structure information of medical images; 
    \item We introduce a window-sliding medical prototype learning method to perceive and utilize positional information, which exploits consistent anatomical patterns and spatial information to strengthen the detection ability of anomalies for radiography images; 
    \item We design a new anomaly scoring method by exploring the concentration and contrast characteristics of the medical anomaly map. 
    \item We demonstrate the superiority and efficiency of SP-Mamba on three medical anomaly detection datasets, while maintaining low model parameters and computational complexity.
\end{itemize}

\section{Related Works}
\subsection{Medical Anomaly Detection}
Given the high level of specialized medical knowledge required to annotate medical images, anomaly detection in this field has traditionally relied on supervised approaches. Recently, unsupervised methods have emerged, offering a way to detect anomalies without the need for extensive manual labeling, thereby easing the burden on medical staff. Early efforts in this area utilize GANs to provide weak annotations for anomaly detection in medical images, with models like AnoGAN~\cite{schlegl2017unsupervisedanomalydetectiongenerative} and f-AnoGAN~\cite{SCHLEGL201930} pioneering this approach. Building on these foundations, SALAD~\cite{zhao_anomaly_2021} introduced a hybrid framework that integrates GANs with self-supervised anomaly detection. Xiang et al. further advanced this field with SQUID~\cite{xiang_squid_2023} and its successor, SimSID~\cite{xiang_exploiting_2024}, which combined GANs with knowledge distillation to enhance detection capabilities. These works also incorporated space-aware memory and memory queues to leverage the structural features of radiographic imaging. Recently, CLIP-based approaches such as MVFA-AD~\cite{Huang_2024_CVPR} have also gained popularity for medical anomaly detection. However, these methods typically require time-consuming pre-training on large medical datasets of image–text pairs and impose heavy computational demands. Meanwhile, Lu et al. proposed Hetero-AE~\cite{lu_anomaly_2024}, a hybrid CNN-Transformer network that balances the modeling of long-range feature dependencies with computational efficiency. This work highlights the growing recognition among researchers of the need to improve computational efficiency in anomaly detection, especially in settings with large data volumes and limited computational resources, such as grassroots hospitals.

Leveraging the unique spatial correlation and shape consistency of medical images, we propose a method that significantly enhances anomaly detection performance. Additionally, we incorporate Mamba, known for its high computational efficiency, as the basis of our anomaly detection framework. This allows us to handle large volumes of medical image data with lower computational costs, making our approach both effective and practical for real-world applications.

\subsection{State Space Models}
Transformers have dominated many deep learning applications thanks to their attention mechanism, which focuses on relevant input parts and captures global context. However, this comes at the cost of high computational demands, especially when dealing with large-scale data. Recently, state space models (SSMs) have emerged as a competitive alternative, offering lower costs and strong capabilities for capturing sequential dependencies.
Mamba~\cite{gu2024mambalineartimesequencemodeling}, a leading variant of SSMs, matches the capabilities of Transformers while maintaining linear scalability with sequence length, inspiring a series of outstanding works in image processing. Vim~\cite{zhu_vision_2024} first introduced Mamba to this field, and VMamba~\cite{liu_vmamba_2024} later proposed a cross-scan module to enhance selective scanning directions in 2D images. Mamba has been particularly impactful in medical image processing. Mamba-UNet~\cite{wang_mamba-unet_2024} combines Mamba blocks with a UNet-like architecture, achieving exceptional performance in medical image segmentation. In image restoration, VmambaIR's~\cite{shi2024vmambairvisualstatespace} omni-selective scan mechanism models image information flow in all six directions, overcoming the unidirectional limitations of traditional SSMs. MambaIRv2~\cite{guo_mambairv2_2025} introduces an attentive state-space equation and a semantic-guided neighboring mechanism, achieving state-of-the-art performance.

More recently, MambaAD~\cite{he_mambaad_2024} applied Mamba to anomaly detection but was not optimized for medical images, which focus on typical photographic imaging and neglect the characteristics of medical radiography imaging. To better fit for medical anomaly detection tasks, we propose a task-oriented model, which taking the radiography imaging into consideration to efficiently focuses on anomalous information in medical images.

\section{Methodology}
\subsection{Overview}
Our proposed model is a Mamba-based reconstruction method that assumes that when trained only with normal images, normal images can be reconstructed accurately during testing, while abnormal images lead to large reconstruction errors. We denote the set of normal images available at training time as $\mathcal{X}_N$ ($\forall{x} \in \mathcal{X}_N: y_{x} = 0$), with $y_{x} \in \{0, 1\}$ denoting if an image $x$ is normal (0) or abnormal (1). Accordingly, we define the test sample as $\forall{x} \in \mathcal{X}_T: y_{x} = \{0,1\}$, including both the normal test images and abnormal test images. 
\begin{figure*}[t!]
    \centering
    \includegraphics[width=1.5\columnwidth]{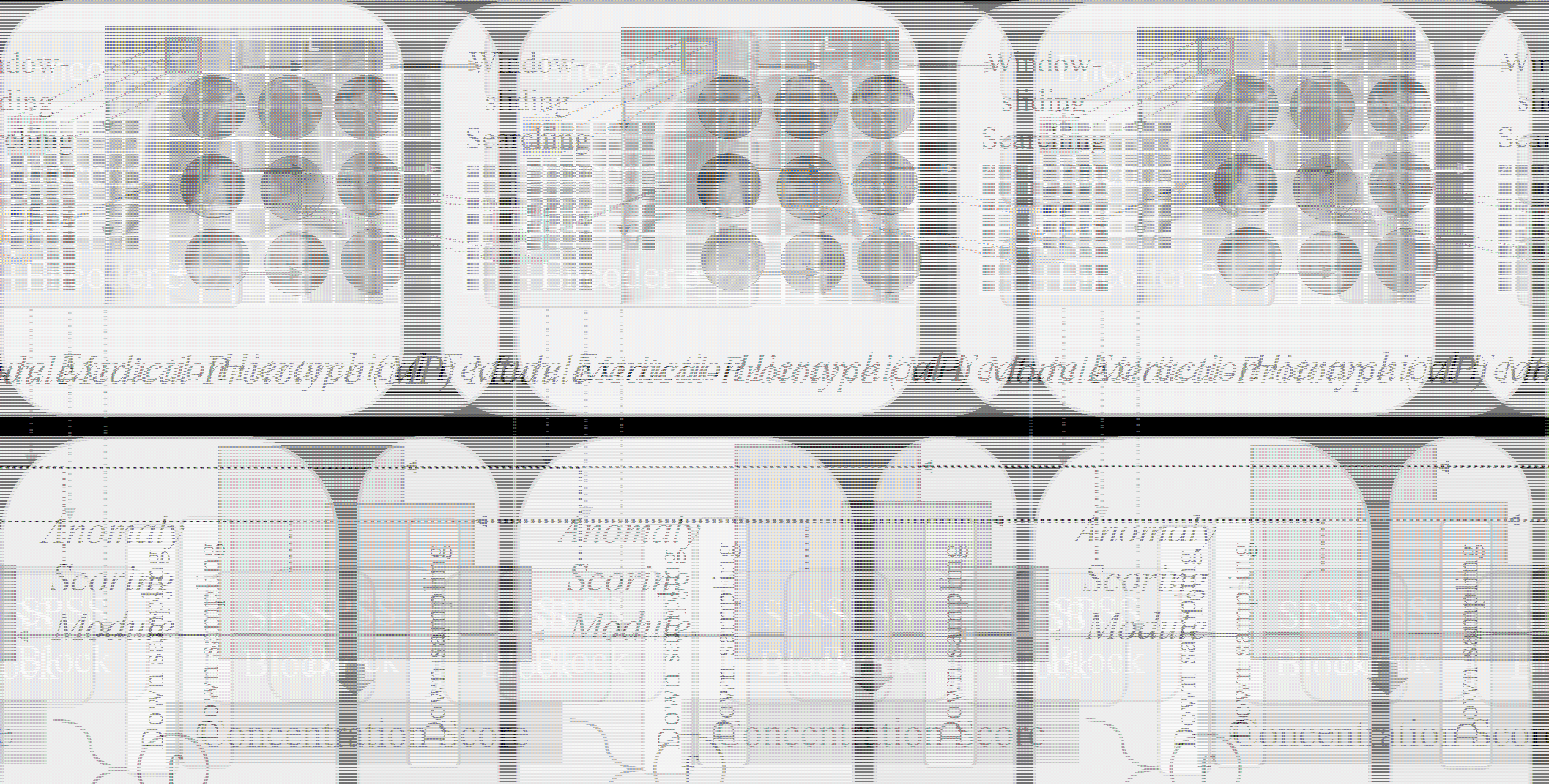}
    \caption{Overview of SP-Mamba: a pyramidal auto-encoder framework for reconstructing multi-scale features while integrating spatial information from medical images. Firstly, the input image is encoded with a pretrained encoder at three scales; Then, the window-sliding medical prototypes are learned to spatial-wise exploit consistent anatomical patterns in the fusion feature space; Next, the hierarchical Mamba-based decoder, SPSS module, is designed with C-H scanning for better leverage position information; Finally, the anomaly score detect and localize anomalies with four components for medical imaging. In total, the spatial information is perceived and leveraged through our whole pipeline, including the MP module, SPSS module, and anomaly scoring module.}
    \label{pipeline}
\end{figure*}
The model pipeline, shown in Figure~\ref{pipeline}, can be summarized as follows: i) The CNN-based encoder, a pre-trained ResNet34, extracts feature maps at three different scales and inputs them into the Half-FPN bottleneck for fusion; ii) The fused output is then fed into the Medical-Prototype (MP) module to refine spatial information and compute the distance between normal features and the prototype as a part of total anomaly score; iii) After that, the aggregated features are fed into the hierarchical Mamba-based Decoder with a depth configuration of [3,4,6,3], corresponding to the depth of the encoder. Within the Mamba-based decoder, we introduce the Spatial-Perception State Space (SPSS) module. Each SPSS Module consists of two branches: one based on Mamba and the other based on multi-layer convolutions. The former branch is used for flexible global perception of the spatial structure information in medical images with the Circular-Hilbert scanning method, while the latter is used for enhancing the complex detail features of medical images locally; iv) During testing, we employ an anomaly score that integrates reconstruction difference, the distance between normal features and the prototype, and the concentration and contrast characteristics of the anomaly map in medical images, which significantly enhances the performance of our proposed model; v) The overall pipeline is optimized under the loss function composed of the average distance between the normal features and the prototype for each patch and the sum of the Mean Squared Error (MSE) computed across feature maps at three scales. 

\subsection{Learning Reliable Window-sliding Medical Prototypes}
Incorporating memory modules into neural networks has been proven effective in multiple deep learning tasks. In this work, we introduce a Medical-Prototype (MP) module that incorporates a memory module into the Mamba framework. This module is tailored to enhance the learning of normal features by prototypes, considering the unique characteristics of medical images. Additionally, the distance between normal features and the prototype is designed as a crucial component of the total anomaly score.

Despite the structural similarities in medical images, the specific body parts in corresponding regions of each image may differ due to variations in body structures, shooting angles, and other factors, which pose challenges for prototypes to learn normal features at corresponding positions. 
\begin{figure}[t!]
    \centering
    \begin{subfigure}{0.45\columnwidth}
        \centering
        \includegraphics[width=0.6\textwidth]{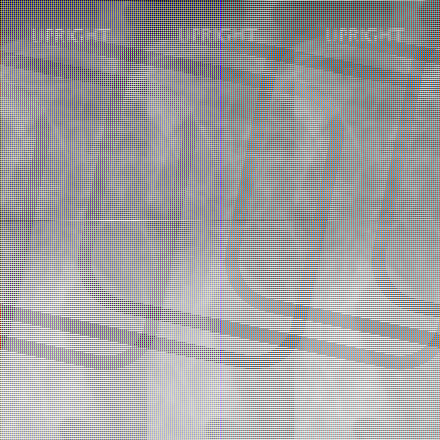}
        \caption{No.00822 from CheXpert}
        \label{fig:No.00822}
    \end{subfigure}
    \hfill
    \begin{subfigure}{0.45\columnwidth}
        \centering
        \includegraphics[width=0.6\textwidth]{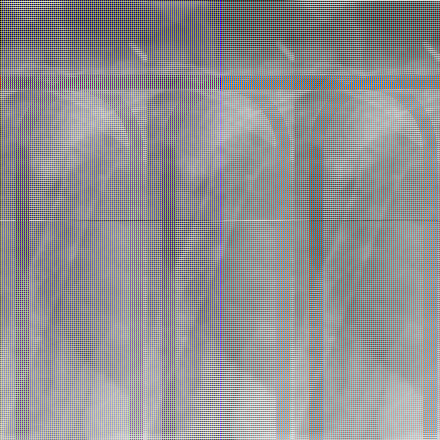}
        \caption{No.00862 from CheXpert}
        \label{fig:No.00862}
    \end{subfigure}
    \caption{Abnormal Images from CheXpert. The specific body parts in corresponding regions of various images may differ.}
    \label{fig:Abnormal Images from CheXpert}
\end{figure}
As shown in Figure~\ref{fig:Abnormal Images from CheXpert}, the patient in~\ref{fig:No.00822} may be affected by scoliosis, causing the lung region to appear tilted in the image, while the patient in~\ref{fig:No.00862} was positioned to the right side of the image during capture, resulting in the lung region also being located on the right side of the image. To address this issue, we introduce a window-sliding selection mechanism to help prototypes match more reliable normal features. Specifically, we perform similarity matching between prototypes and normal features at the patch level, which can effectively preserve the spatial structure information contained within each patch.

Specifically, as shown in Figure~\ref{pipeline}, during the matching process, we do not only calculate the cosine similarity between each patch of the prototype and the corresponding position patch of the normal feature. Instead, we calculate the cosine similarity between each patch of the prototype and every patch within a $p\times p$ range centered on the corresponding position patch of the normal feature. We ultimately select the minimum cosine distance between the prototype's corresponding patch and all patches within the sliding window range, incorporating this distance into the distance between the normal features. This allows the prototype to learn more accurate and reliable normal features. During the matching process, the $p\times p$ window slides with the target patch of the prototype. When the target patch of the prototype is located at the corners, the window only includes $\frac{(p+1)^{2}}{4}$ patches. When it is on the edge but not at a corner, the window includes $\frac{p(p+1)}{2}$ patches. When it is in the center, the window encompasses all the $p^{2}$ patches. The varying number $N$ of patches within the window for different target patch positions does not weaken the prototype's learning ability. Instead, due to the secondary and random nature of spatial information at the edges of medical images, this approach effectively enhances the robustness of the prototype's learning. 

Under the trade-off between anomaly detection performance and computational efficiency on different datasets, we set different numbers $K$ of prototypes. After each prototype learns the normal features, we extract the patch with the highest similarity to the normal features from the corresponding position patches of each prototype to form the final prototype. The distance $\mathcal{D}$ between the normal features and the final prototype can be calculated using the following formula:
\begin{equation}\label{eqn-1}
\begin{aligned}
M_K^r &=\frac{m_K^* \cdot m_K^r}{ \left \|m_K^* \right \| \times \ \left \|m_K^r\right \|},\\
D_K &=1 - max(M_K^1,M_K^2,\ldots,M_K^N ),\\
\mathcal{D} &= \min\left(D_1, D_2, \ldots, D_K\right),\
\end{aligned}
\end{equation}

where $m_K^*$ represents the target patch in the K-th prototype, $m_K^i$ represents the patch in the sliding window that is used to compute the similarity with the prototype's patch, $M_K^i$ represents the cosine similarity between the two, and $D_K$ represents the distance between the K-th prototype and the normal features. The distance $\mathcal{D}$ between normal features and the final prototype is then upsampled to the same shape of $H \times W$  as the anomaly map for calibrating the anomaly score, and together with the sum of the Mean Squared Error (MSE) computed across feature maps at three scales, constitutes the total loss function as:
\begin{equation}\label{eqn-2}
 \mathbb{L} =\sum_{l=1}^{L} \frac{1}{n} \left \|f^{l}_{org}-f^{l}_{rec}  \right \| ^{2}_{2}+\varepsilon\frac{1}{H\cdot W}\sum_{i=1}^{H}\sum_{j=1}^{W}\mathcal{D}_{i,j},\
\end{equation}

where $\varepsilon$ is the weighting coefficient for the weighted sum.

This method optimizes the prototype and further normalizes the learning of reconstructed features from normal features.
\subsection{Circular-Hilbert Scanning Method for Mamba}

Mamba~\cite{gu2024mambalineartimesequencemodeling} scans context using an input-dependent selection mechanism (referred to as S6), and VMamba~\cite{liu_vmamba_2024} proposes 2D-Selective-Scan, which rearranges the input patches into sequences by cross-scanning in four different directions for visual tasks. Traditional scanning methods' equal attention to all patches can lead to interference from unrelated patches, while other unrelated positions are filled with distracting information, making it difficult to focus correctly. To address this issue of misdirected attention, we propose the Circular-Hilbert scanning method, which considers the determinacy and concentration of lesion locations in medical images.
\begin{figure}[t!]
    \centering
    \includegraphics[width=\columnwidth]{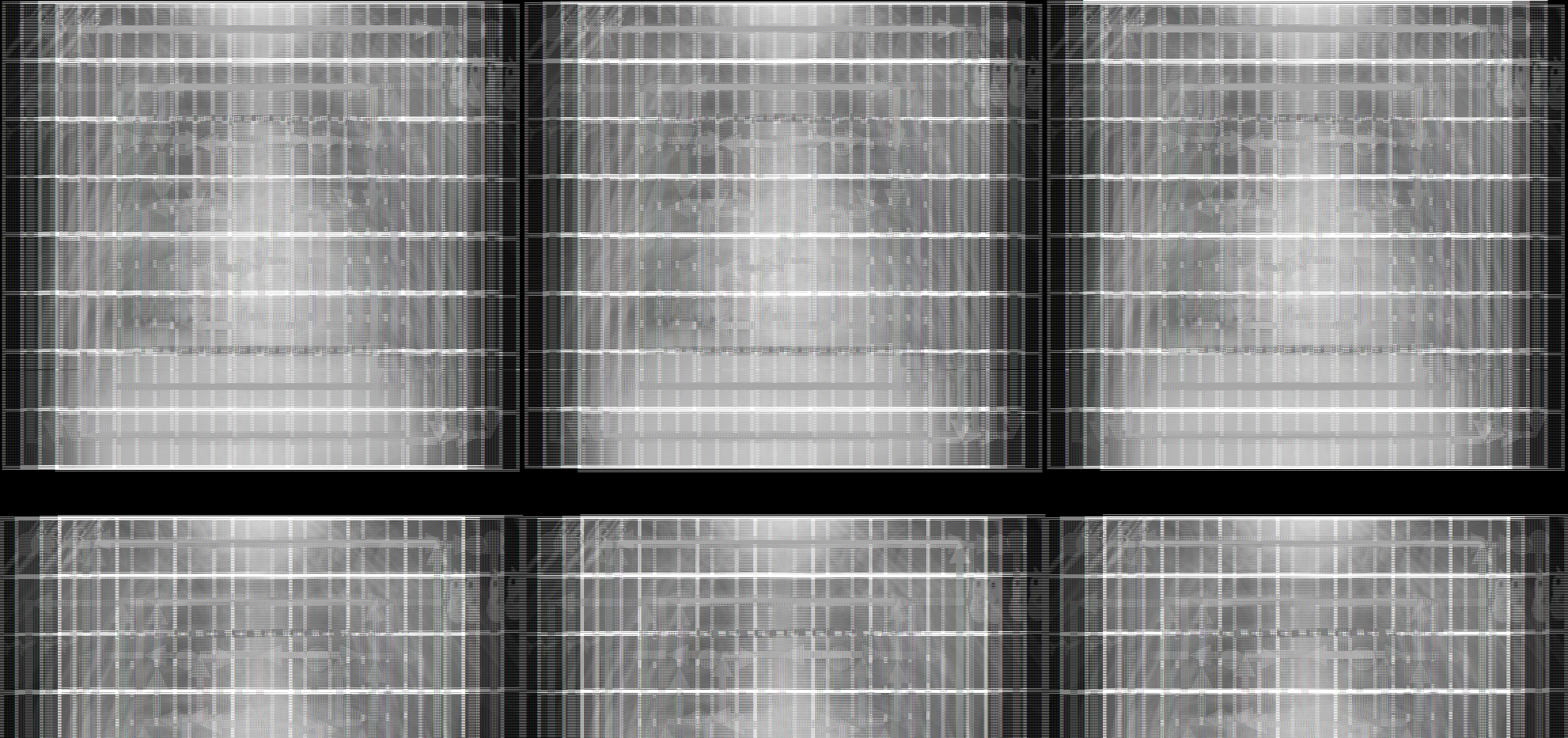}
    \caption{Circular-Hilbert Scan, containing 8 scanning directions used for both C-H Encoder and Decoder.}
    \label{C-H Scan}
    \vspace{-4mm}
\end{figure}
As shown in Figure~\ref{C-H Scan}, we utilize the Circular-Hilbert scanning method in every Spatial-Perception State Space (SPSS) block. The specific description of the Circular-Hilbert scanning method is as follows: For each image feature input to the SPSS block, we divide it into $h'\times w'$ patches. For the central $\frac{h'}{2}\times \frac{w'}{2}$ patches, we apply the Hilbert scanning method, while for the surrounding patches, we implement the circular scanning method. The central Hilbert curve can be obtained by an n-order Hilbert matrix:
\begin{equation}\label{eqn-7}
\begin{aligned}
H_{n+1} = \left\{ \begin{array}{ll}
\begin{pmatrix}
H_n & 4^n E_n + H_n^T \\
(4^{n+1} + 1) E_n - H_n^{ud} & (3 \times 4^n + 1) E_n - (H_n^{lr})^T
\end{pmatrix},\\ 
\qquad\qquad\qquad\qquad\qquad\qquad\qquad\qquad\text{if } n \text{ is even}. \\
\begin{pmatrix}
H_n & (4^{n+1} + 1) E_n - (H_n^T)^{lr} \\
4^n E_n + H_n^T & (3 \times 4^n + 1) E_n - H_n^{ud}
\end{pmatrix}, \\
\qquad\qquad\qquad\qquad\qquad\qquad\qquad\qquad\text{if } n \text{ is odd}.
\end{array} \right.
\end{aligned}
\end{equation}

The central Hilbert curve maintains the locality of points in 2D space, ensuring adjacent points remain adjacent on the curve. This characteristic allows for maximal retention of spatial correlation when mapping a 2D image to 1D. In contrast, the outer circular scanning method cannot fully leverage the positional relationships between adjacent patches. By reducing the focus on the outer regions, it mitigates the interference from areas unrelated to the lesion in anomaly detection. Furthermore, the Circular-Hilbert scanning method can model image features from a maximum of 16 directions. These 16 directions are determined by the following scanning details: clockwise and counterclockwise scanning using the outer circular scanning method, using the 4 corner patches as starting or ending positions for scanning, and scanning from the outer circle toward the center or from the center toward the outer circle. Theoretically, this can extend to more spatially informative sequences when modeling extremely large medical datasets used in practical applications in hospitals, allowing for a more comprehensive and effective utilization of 2D image spatial information, thereby enhancing the encoding and modeling capabilities of feature sequences.

In summary, our proposed Circular-Hilbert scanning method can fully exploit the structural similarity of medical images and the concentrated distribution characteristics of lesions. Meanwhile, as the modeling capability of the outer Circular scanning method is inferior to that of the central Hilbert scanning method, the Circular-Hilbert scanning method effectively mitigates the negative impact of structural randomness in the image edge regions on modeling accuracy, which is caused by various factors such as variations in shooting angles.

\subsection{Concentration and Contrast Characteristics of Anomaly Maps in Medical Images}
\label{section:3.4}
We notice that there are significant differences between the anomaly maps of normal medical images and those of abnormal images. This difference can also be particularly leveraged for medical anomaly detection.
\begin{figure}[t!]
    \centering
    \begin{subfigure}{0.45\columnwidth}
        \centering
        \includegraphics[width=0.6\textwidth]{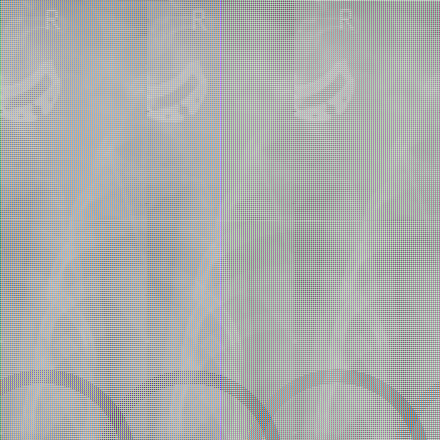}
        \caption{Visualized Anomaly Map of the Abnormal Image}
        \label{fig:Visualized Anomaly Map of the Abnormal Image}
    \end{subfigure}
    \hfill
    \begin{subfigure}{0.45\columnwidth}
        \centering
        \includegraphics[width=0.6\textwidth]{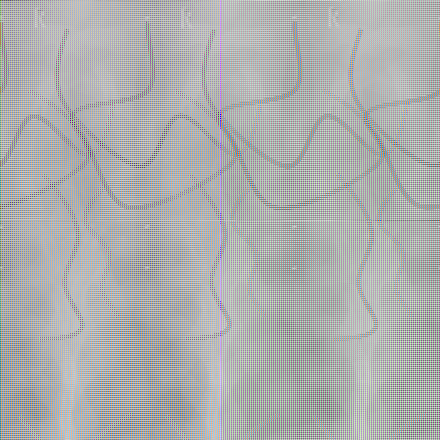}
        \caption{Visualized Anomaly Map of the Normal Image}
        \label{fig:Visualized Anomaly Map of the Normal Image}
    \end{subfigure}
    \caption{Visualized Anomaly Maps from ZhangLab Chest X-ray. The anomaly maps of abnormal images show concentrated characteristics, while the anomaly maps of normal images are more dispersed.}
    \label{fig:Visualized Anomaly Maps from Zhanglab}
    \vspace{-5mm}
\end{figure}

\textbf{Concentration Score.} \quad As shown in Figure~\ref{fig:Visualized Anomaly Maps from Zhanglab}, the visualization results of the anomaly map for medical images show that both abnormal and normal images have certain highlights in their anomaly maps.  However, the key difference lies in the distribution of these highlights. For abnormal images, the highlights in the anomaly map are more concentrated, clearly indicating the location of the lesion. In contrast, for normal images, the highlights are more dispersed and do not pinpoint the location of any lesion. This observation reflects that the information distribution in the anomaly map of abnormal medical images is more concentrated, while that in normal medical images is more dispersed. Based on this difference, we can design an anomaly score with greater discriminability. For anomaly maps with a shape of $H\times W$, our $S_{concen}$, anomaly score designed based on the information concentration characteristic of the anomaly map in medical images, is specifically calculated through the following algorithm:
\begin{equation}\label{eqn-8}
 S_{concen} = \frac{1}{H \cdot W} \sum_{i=1}^{H} \sum_{j=1}^{W} C_{i,j}d_{i,j},\
\end{equation}

where the cost score $C_{i,j}$ is defined as $C_{i,j} = (e_{i,j}-e^{*})^{2}$ and the distance $d_{i,j}$ is defined as $d_{i,j} = \sqrt{(x_{i,j}-x^{*})^{2}+(y_{i,j}-y^{*})^{2}}$. $e_{i,j}$ and $e^{*}$ respectively represent the value of the pixel at position $(x_{i,j}, y_{i,j})$ on the anomaly map and the maximum value on the anomaly map. Correspondingly, $(x^{*}, y^{*})$ represents the position of the pixel with the maximum value on the anomaly map.

\textbf{Contrast Score.}\quad As shown in Figure~\ref{fig:Visualized Anomaly Maps from Zhanglab}, there is also a significant difference in contrast between the anomaly maps of normal and abnormal medical images. For anomaly maps with a shape of $H\times W$, we can use the difference of Gaussians(DoG) to quantify this contrast characteristic into a contrast score $S_{contra}$, with the specific algorithm as follows: 
\begin{equation}\label{eqn-9}
\begin{aligned}
G(x_{i,j},y_{i,j},\sigma) & =\frac{1}{2\pi \sigma ^{2}} e^{-\frac{x^{2}_{i,j}+y^{2}_{i,j}}{2\sigma ^{2}}}, \\
DoG(x_{i,j},y_{i,j},\sigma, k\sigma) & =G(x_{i,j},y_{i,j},k\sigma)-G(x_{i,j},y_{i,j},\sigma),\\
S_{contra} & =\frac{1}{H \cdot W} \sum_{i=1}^{H} \sum_{j=1}^{W}DoG(x_{i,j},y_{i,j},\sigma, k\sigma),\\
\end{aligned}
\end{equation}

where $G(x_{i,j},y_{i,j},\sigma)$ is a two-dimensional Gaussian function with mean $\sigma$, and $DoG(x_{i,j},y_{i,j},\sigma, k\sigma)$ is the Difference of Gaussians (DoG) formula with parameters $x_{i,j}$, $y_{i,j}$, $\sigma$, and $k\sigma$.

\textbf{Prototype Score.}\quad Accordingly, in our proposed method, we note that the anomaly degree could be reflected by the dissimilarity between normal features and prototypes. Therefore, the distance $\mathcal{D}$ calculated by formula~\ref{eqn-1}, which captures the significant differences between the prototype and normal features at anomalous regions, can also serve as a part of the total anomaly score $S_{p-dist}$ and play a crucial role in calibrating the anomaly score. 

\textbf{Anomaly Scoring.}\quad We calculate the reconstruction difference by summing the cosine similarity distances between the normal features and the reconstructed features at various scales, as specifically shown below:
\begin{equation}\label{eqn-10}
 S_{org}=\sum_{l=1}^{L} 1-\frac{f^{l}_{org} \cdot f^{l}_{rec}}{ \left \|f^{l}_{org} \right \| \times \ \left \|f^{l}_{rec}\right \|} .\
\end{equation}

In our proposed method, we notice that the anomaly degree could also be reflected by the dissimilarity between normal features and prototypes. Therefore, the distance $\mathcal{D}$, which captures the significant differences between the prototype and normal features at anomalous regions, can also serve as a part of the total anomaly score $S_{p-dist}$ and play a crucial role in Calibrating the Anomaly Score.
During the test phase, the total anomaly score $S_{total}$ is composed of four parts: reconstruction difference as $S_{org}$, the distance between normal features and the prototype as $S_{p-dist}$, concentration score as $S_{concen}$, and contrast score as $S_{contra}$.The calculation formula for $S_{total}$ is as follows:
\begin{equation}\label{eqn-11}
 S_{total} = S_{org}+\alpha S_{p-dist}+\beta S_{concen}+\gamma S_{contra}.
\end{equation}

\section{Experiments}
\subsection{Experimental Setting}

\textbf{Datasets:} We evaluate the performance of our model on multimodal medical datasets, including chest X-ray images, brain MRI images, liver CT images, and retinal OCT images.

\begin{table*}[t!]
\centering
\caption{Benchmark Results(AUC/Acc/F1 Score) on the Test Set of ZhangLab Chest X-ray}
\label{tab:zhanglab}
\setlength{\tabcolsep}{25pt}
\begin{tabular}{@{}lccccc@{}}
\toprule
\quad Method & Ref \& Year & AUC (\%) & Acc (\%) & F1 (\%) \\ \midrule
\quad SALAD~\cite{9469869} & TMI'21 & 82.7$\pm$0.8 & 75.9$\pm$0.9 & 82.1$\pm$0.3 \\
\quad CutPaste~\cite{Li_2021_CVPR} & CVPR'21 & 73.6$\pm$3.9 & 64.0$\pm$6.5 & 72.3$\pm$8.9 \\
\quad PANDA~\cite{Reiss_2021_CVPR} & CVPR'21 & 65.7$\pm$1.3 & 65.4$\pm$1.9 & 66.3$\pm$1.2 \\
\quad M-KD~\cite{Salehi_2021_CVPR} & CVPR'21 & 74.1$\pm$2.6 & 69.1$\pm$0.2 & 62.3$\pm$8.4 \\
\quad IF 2D~\cite{marimont2021implicitfieldlearningunsupervised} & MICCAI'21 & 81.0$\pm$2.8 & 67.4$\pm$0.2 & 82.2$\pm$2.7 \\
\quad PaDiM~\cite{defard2020padimpatchdistributionmodeling} & ICPR'21 & 71.4$\pm$3.4 & 72.9$\pm$2.4 & 80.7$\pm$1.2 \\
\quad IGD~\cite{chen2022deeponeclassclassificationinterpolated} & AAAI'22 & 73.4$\pm$1.9 & 74.0$\pm$2.2 & 80.9$\pm$1.3 \\
\quad SQUID~\cite{xiang_squid_2023} & CVPR'23 & 87.6$\pm$1.5 & 80.3$\pm$1.3 & 84.7$\pm$0.8 \\
\quad SimSID~\cite{xiang_exploiting_2024} & TPAMI'24 & \underline{91.1$\pm$0.9} & \underline{85.0$\pm$1.0} & \underline{88.0$\pm$1.1} \\ 
\quad MambaAD~\cite{he_mambaad_2024} & NeurIPS'24 & 86.2$\pm$4.6 & 81.3$\pm$2.2 & 86.0$\pm$2.3 \\ \bottomrule
\quad SP-Mamba & Ours & \textbf{92.0$\pm$2.4} & \textbf{86.0$\pm$0.9} & \textbf{88.9$\pm$0.8}
\end{tabular}
\end{table*}

\begin{table*}[t!]
\centering
\caption{Benchmark Results(AUC/Acc/F1 Score) on the Test Set of CheXpert}
\label{tab:cheXpert}
\setlength{\tabcolsep}{28pt}
\begin{tabular}{@{}lccccc@{}}
\toprule
\quad Method & Ref \& Year & AUC (\%) & Acc (\%) & F1 (\%) \\ \midrule
\quad CutPaste~\cite{Li_2021_CVPR} & CVPR'21 & 65.5$\pm$2.2 & 62.7$\pm$2.0 & 60.3$\pm$4.6 \\
\quad PANDA~\cite{Reiss_2021_CVPR} & CVPR'21 & 68.6$\pm$0.9 & 66.4$\pm$2.8 & 65.3$\pm$1.5 \\
\quad M-KD~\cite{Salehi_2021_CVPR} & CVPR'21 & 69.8$\pm$1.6 & 66.0$\pm$2.5 & 63.6$\pm$5.7 \\
\quad SQUID~\cite{xiang_squid_2023} & CVPR'23 & 78.1$\pm$5.1 & 71.9$\pm$3.8 & \textbf{75.9$\pm$5.7} \\
\quad SimSID~\cite{xiang_exploiting_2024} & TPAMI'24 & \underline{79.7$\pm$2.2} & \underline{72.9$\pm$1.9} & 71.9$\pm$2.3 \\
\quad MambaAD~\cite{he_mambaad_2024} & NeurIPS'24 & 73.1$\pm$3.1 & 68.4$\pm$2.5 & 72.3$\pm$2.2 \\ \bottomrule
\quad SP-Mamba & Ours & \textbf{80.4$\pm$0.9} & \textbf{74.8$\pm$1.2} & \underline{75.8$\pm$1.5}
\end{tabular}
\end{table*}

\begin{table*}[ht!]
\centering
\caption{Benchmark Results(AUC/AP/F1 Score) on the Test Set of Uni-Medical}
\label{tab:uni-medical}
\scalebox{0.9}{
\begin{tabular}{ccccc}
\hline
Method$\rightarrow$ & DeSTSeg(CVPR'23)~\cite{Zhang_2023_CVPR} & DiAD(AAAI'24)~\cite{he2023diaddiffusionbasedframeworkmulticlass} & MambaAD(NeurIPS'24)~\cite{he_mambaad_2024} & SP-Mamba(Ours)  \\ \cline{1-1}
Category$\downarrow$ & image-level\qquad pixel-level        & image-level\qquad pixel-level        & image-level\qquad pixel-level     & image-level\qquad pixel-level \\ \hline
brain    & 84.5/95.0/92.1\quad89.3/33.0/37.0 & 93.7/98.1/\textbf{95.0}\quad95.4/42.9/36.7 & \underline{94.2}/\underline{98.6}/\underline{94.5}\quad\underline{98.1}/\underline{62.7}/\underline{62.2} & \textbf{94.3}/\textbf{98.7}/94.4\quad\textbf{98.2}/\textbf{65.8}/\textbf{65.1}       \\
liver    & \underline{69.2}/\textbf{60.6}/\underline{64.7}\quad79.4/\textbf{21.9}/\textbf{28.5} & 59.2/55.6/60.9\quad\underline{97.1}/\underline{13.7}/7.3 & 63.2/53.1/\underline{64.7}\quad96.9/9.1/16.3 & \textbf{70.2}/\underline{58.7}/\textbf{67.9}\quad\textbf{97.2}/10.8/\underline{18.5}       \\
retinal  & 88.3/83.8/79.2\quad91.0/59.0/46.8 & 88.3/86.6/77.7\quad95.3/57.5/62.8 & \underline{93.6}/\underline{88.7}/\underline{86.6}\quad\underline{95.7}/\underline{64.5}/\underline{63.4} & \textbf{95.7}/\textbf{91.3}/\textbf{87.8}\quad\textbf{96.1}/\textbf{66.9}/\textbf{64.8}       \\
Mean     & 80.7/79.8/78.7\quad86.6/38.0/37.5 & 80.4/\underline{80.1}/77.8\quad95.9/38.0/35.6 & \underline{83.7}/\underline{80.1}/\underline{82.0}\quad\underline{96.9}/\underline{45.4}/\underline{47.3} & \textbf{86.7}/\textbf{82.9}/\textbf{83.4}\quad\textbf{97.2}/\textbf{47.8}/\textbf{49.5}       \\ \hline
\end{tabular}
}    
\end{table*}

\texttt{ZhangLab Chest X-ray~\cite{kermany_identifying_2018}.}\quad Released by Zhang Lab at the University of San Diego, USA, this dataset includes healthy (normal) and pneumonia (anomaly) images and has been widely used in medical anomaly detection and classification tasks. Following the approach of Xiang et al.~\cite{xiang_exploiting_2024}, we use 1249 healthy images from the official training set for training and a test set comprising 234 normal and 390 abnormal images.

\texttt{CheXpert~\cite{irvin2019chexpertlargechestradiograph}.}\quad The dataset released by Andrew Ng's team at Stanford University in 2019 is a large dataset of chest X-rays and competition for automated chest X-ray interpretation. Following Xiang et al.~\cite{xiang_squid_2023}, we utilize 4499 healthy images from the official training set for training and a balanced test set of 250 normal and 250 abnormal images.

\texttt{Uni-Medical~\cite{he_mambaad_2024}.}\quad He et al.~\cite{he_mambaad_2024} carefully select three benchmarks from BMAD~\cite{bao_bmad_2024} and integrate them into a multimodal dataset including three classes: brain, liver, and retinal. The brain class contains 7,500 images for training and 3,715 images for testing. The liver class consists of 1,542 images for training and 1,493 images for testing. The retinal class includes 4,297 images for training and 1,805 images for testing.

\textbf{Metrics:} Following Xiang et al.~\cite{xiang_squid_2023}, we evaluate the performance using three metrics for anomaly detection: Area Under the Receiver Operating Characteristic Curve (AUC), Accuracy(Acc), and F1 Score on the ZhangLab Chest X-Ray and Stanford CheXpert datasets. Moreover, we report Area Under the Receiver Operating Characteristic Curve (AUC), Average Precision(AP) and F1 Score on the Uni-Medical dataset for anomaly detection and localization, following the approach of MambaAD~\cite{he_mambaad_2024}. To compare efficiency, we utilize two metrics: Params and FLOPs, following MambaAD~\cite{he_mambaad_2024}. Additionally, we represent the average values of image-level AUC, AP and F1 Score as mAD.

\textbf{Implementation Details:} All input images are resized to a uniform size of $256\times 256$ before extracting their features. A pre-trained ResNet34 with the depth of [3,4,6,3] is utilized as the feature extractor, and the numbers of the Mamba blocks in each SPSS block corresponds to it. To improve computational efficiency, we train and test using 8 scanning directions of the Circular-Hilbert scanning method, as shown in Figure~\ref{C-H Scan}. These directions are differentiated based on the following details: using patches from the upper left and upper right corners of the image feature as the starting or ending scanning positions; applying the outer circular scanning method in both clockwise and counterclockwise directions; and distinguishing between scanning from the outer circle toward the center and scanning from the center outward. 
During the training period, the AdamW optimizer is employed with an initial learning rate of 0.005 and a decay rate of $1\times10^{-4}$, and all training processes are conducted on a single NVIDIA GeForce RTX 3090 24GB GPU with a batch size of 16. During training, the weight $\varepsilon$ we select for incorporating $\mathcal{D}$ into the total loss function is 25, effectively normalizing the training process.

\subsection{Comparison with SOTA Methods}
We compare our SP-Mamba on three medical datasets: ZhangLab Chest X-ray, CheXpert, and Uni-Medical, against other methods, especially the SOTAs in medical anomaly detection: SQUID and SimSID, as well as MambaAD based on Mamba. The comparisons show that our SP-Mamba demonstrates strong anomaly detection capabilities across multimodal medical datasets, including chest X-ray images, brain MRI images, liver CT images, and retinal OCT images, contributing to the advancement of multimodal medical imaging lesion detection technology.

\textbf{Quantitative Results.}\quad As shown in Table~\ref{tab:zhanglab} and~\ref{tab:cheXpert}, our SP-Mamba outperforms all the comparative methods and reaches a new SOTA in medical anomaly detection on the ZhangLab Chest X-ray dataset and CheXpert dataset. Additionally, as shown in Table~\ref{tab:uni-medical}, on the multi-class medical dataset, Uni-Medical dataset, our SP-Mamba continues to perform well on both anomaly detection and localization, achieving a new SOTA result. To demonstrate the superiority of our model in terms of efficiency, we compare our SP-Mamba with other SOTA methods, as shown in Table~\ref{tab:efficiency_1} and Table~\ref{tab:efficiency_2} .

\textbf{Qualitative Results.}\quad We conduct qualitative experiments on three datasets: ZhangLab Chest X-ray, CheXpert, and Uni-Medical, which qualitatively demonstrate the accuracy of our model in localizing lesions in medical images, as shown in Figure~\ref{fig:Qualitative Results}. Moreover, more qualitative results can be found in the appendix and Figures ~\ref{brain} to ~\ref{retinal} demonstrate the results of our SP-Mamba in comparison with MambaAD~\cite{he_mambaad_2024} in terms of anomaly localization. The qualitative experiments on multimodal medical datasets also reflect the broad prospects of our proposed SP-Mamba in multimodal applications. 

Both quantitative and qualitative results demonstrate the superior performance of our SP-Mamba in detecting and localizing lesions in medical images. This can be attributed to our particular design for exploiting spatial information in medical images. Specifically, the Circular-Hilbert scanning method efficiently scans concentrated lesion areas, the Medical-Prototype module effectively better exploits consistent anatomical patterns and leverages spatial information, and $S_{concen}$ and $S_{contra}$ could model the concentration and contrast characteristics for better anomaly detection of medical images.

\subsection{Ablation Study}
We conduct ablation studies to validate the practical effectiveness of each component in our framework. Taking MambaAD~\cite{he_mambaad_2024} as the baseline, we incrementally incorporate the Circular-Hilbert scanning method, the Medical-Prototype Module, $S_{contra}$ and $S_{concen}$ in our experiments. Table~\ref{tab:ablation} demonstrates the effectiveness of each component of our model, with the best performance achieved by the full version of SP-Mamba.

Compared to the baseline, the version with only the Circular-Hilbert scanning method performs better in most metrics, showing its effectiveness in focusing on lesion information in medical images. Comparing this version with the one further incorporating the Medical-Prototype Module highlights the latter's importance in using consistent anatomical patterns and spatial information for medical anomaly detection. The full model outperforms the version without $S_{concen}$ and $S_{contra}$ in all metrics, demonstrating the value of fully utilizing the medical anomaly map characteristics embodied by $S_{concen}$ and $S_{contra}$.

All the above components revolve around innovations that leverage the spatial information in medical images. By fully exploiting this spatial knowledge during both training and inference, our SP-Mamba establishes a new state-of-the-art for medical image anomaly detection.

\begin{table*}[ht!]
\centering
\caption{Ablation Study on ZhangLab Chest X-ray and CheXpert}
\label{tab:ablation}
\begin{tabular}{cccc|ccc|ccc}
\hline
\multirow{2}{*}{C-H Scan} & \multirow{2}{*}{Prototype} & \multirow{2}{*}{$S_{contra}$} & \multirow{2}{*}{$S_{concen}$} & \multicolumn{3}{c|}{Zhanglab} & \multicolumn{3}{c}{CheXpert}\\ 
 & & & & \multicolumn{1}{c}{AUC(\%)} & \multicolumn{1}{c}{Acc(\%)} & \multicolumn{1}{c|}{F1(\%)} & \multicolumn{1}{c}{AUC(\%)} & \multicolumn{1}{c}{Acc(\%)} & \multicolumn{1}{c}{F1(\%)} \\ \hline
- & - & - & - & 86.2 & 81.3 & 86.0 & 73.1 & 68.4 & 72.3\\
\checkmark & - & - & - & 90.0 & 84.5 & 87.8 & 72.5 & 69.2 & 72.0 \\
\checkmark & \checkmark & - & - & 89.8 & 84.6 & 88.2 & 73.1 & 70.0 & 71.6 \\
\checkmark & \checkmark & \checkmark & - & 90.5 & 85.6 & 88.7 & 79.2 & 74.0 & 75.0 \\
\checkmark & \checkmark & \checkmark & \checkmark & \textbf{92.0} & \textbf{86.0} & \textbf{89.0} & \textbf{80.4} & \textbf{74.8} & \textbf{75.8} \\ \hline
\end{tabular}
\end{table*}

\begin{table}[t!]
\centering
\caption{Efficiency Comparison on ZhangLab Chest X-ray Datasets}
\label{tab:efficiency_1}
\scalebox{0.9}{
\begin{tabular}{cccccc}
\hline
Method    & Params(M) & FLOPs(G) & AUC(\%) & Acc(\%) & F1(\%)   \\ \hline
SQUID~\cite{xiang_squid_2023} & 44.5 & 15.2 & 87.6 & 80.3 & 84.7     \\
SimSID~\cite{xiang_exploiting_2024} & 83.9 & 15.2 & 91.1 & 85.0	& 88.0     \\
SP-Mamba     & \textbf{25.8} & \textbf{8.3} & \textbf{92.0} & \textbf{86.0} & \textbf{88.9}    \\ \hline
\end{tabular}
}
\end{table}

\begin{table}[t!]
\centering
\caption{Efficiency Comparison on Uni-Medical}
\label{tab:efficiency_2}
\begin{tabular}{cccc}
\hline
Method    & Params(M) & FLOPs(G) & mAD \\ \hline
RD4AD~\cite{Deng_2022_CVPR} & 80.6 & 28.4 & 76.5   \\
SimpleNet~\cite{Liu_2023_CVPR} & 72.8 & 16.1 & 76.4   \\
DeSTSeg~\cite{Zhang_2023_CVPR} & 35.2 & 122.7 & 80.0   \\
DiAD~\cite{he2023diaddiffusionbasedframeworkmulticlass} & 1331.3 & 451.5 & 79.4     \\
MambaAD~\cite{he_mambaad_2024} &\textbf{25.7} &\textbf{8.3} & \underline{81.9}     \\
SP-Mamba     & \underline{25.8} & \textbf{8.3} & \textbf{84.3}    \\ \hline
\end{tabular}
\end{table}

\begin{figure}[t!]
    \centering
    \includegraphics[width=\columnwidth]{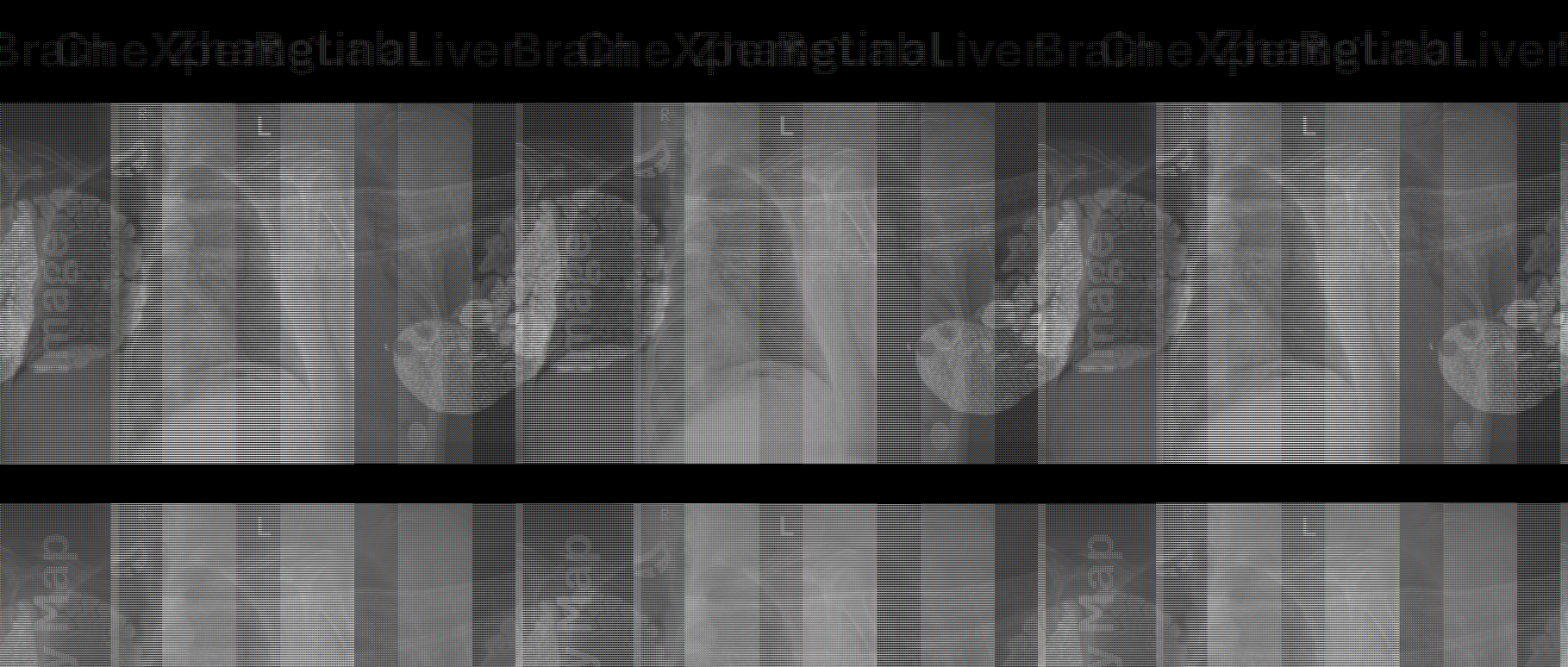}
    \caption{Qualitative Results on various datasets.}
    \label{fig:Qualitative Results}
    \vspace{-3mm}
\end{figure}

\begin{figure}[t!]
    \centering
    \begin{subfigure}{0.45\columnwidth}
        \centering
        \includegraphics[width=0.9\textwidth]{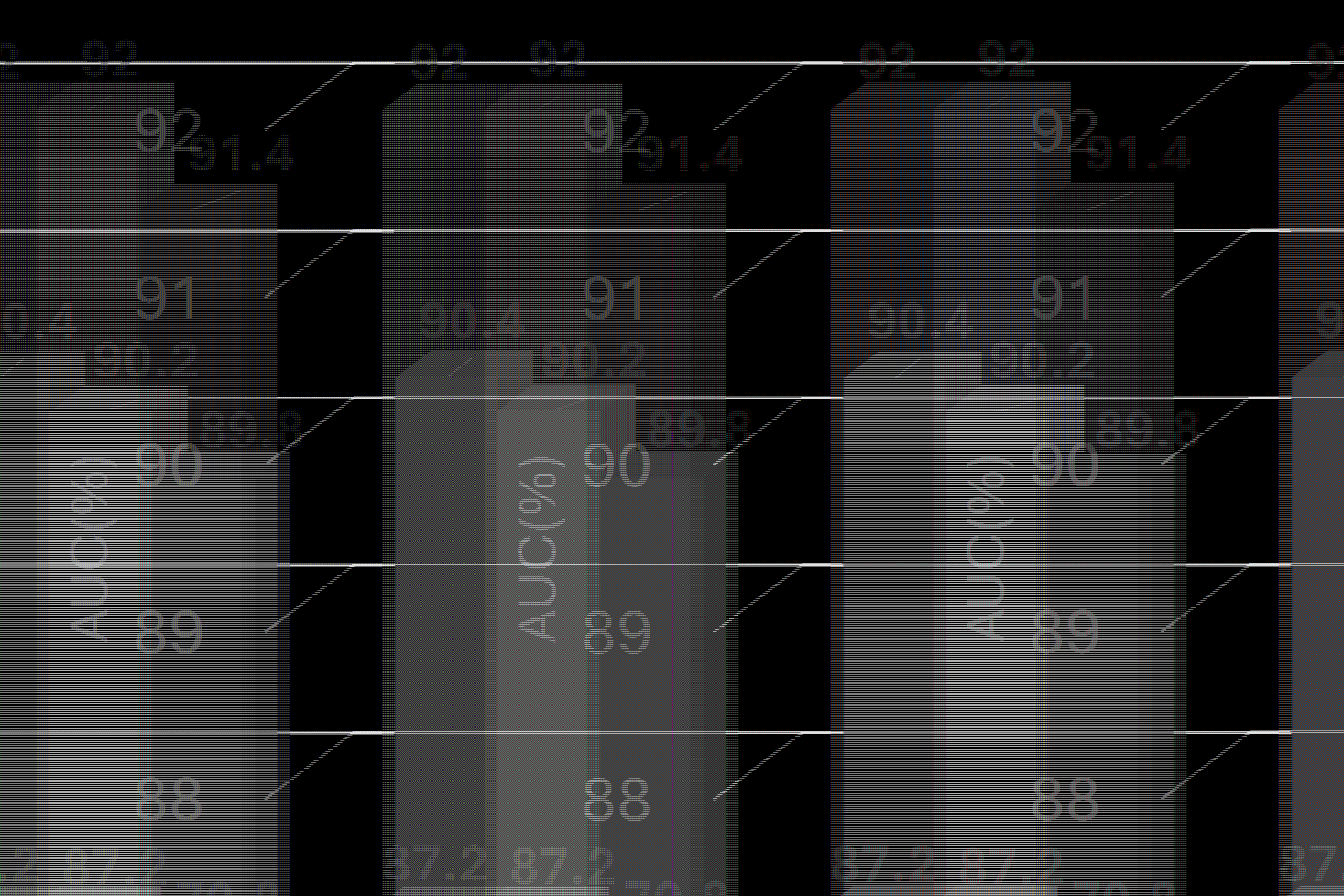}
        \caption{Analysis of $K$ and $\alpha$}
        \label{fig:hyper1}
    \end{subfigure}
    \hfill
    \begin{subfigure}{0.45\columnwidth}
        \centering
        \includegraphics[width=0.9\textwidth]{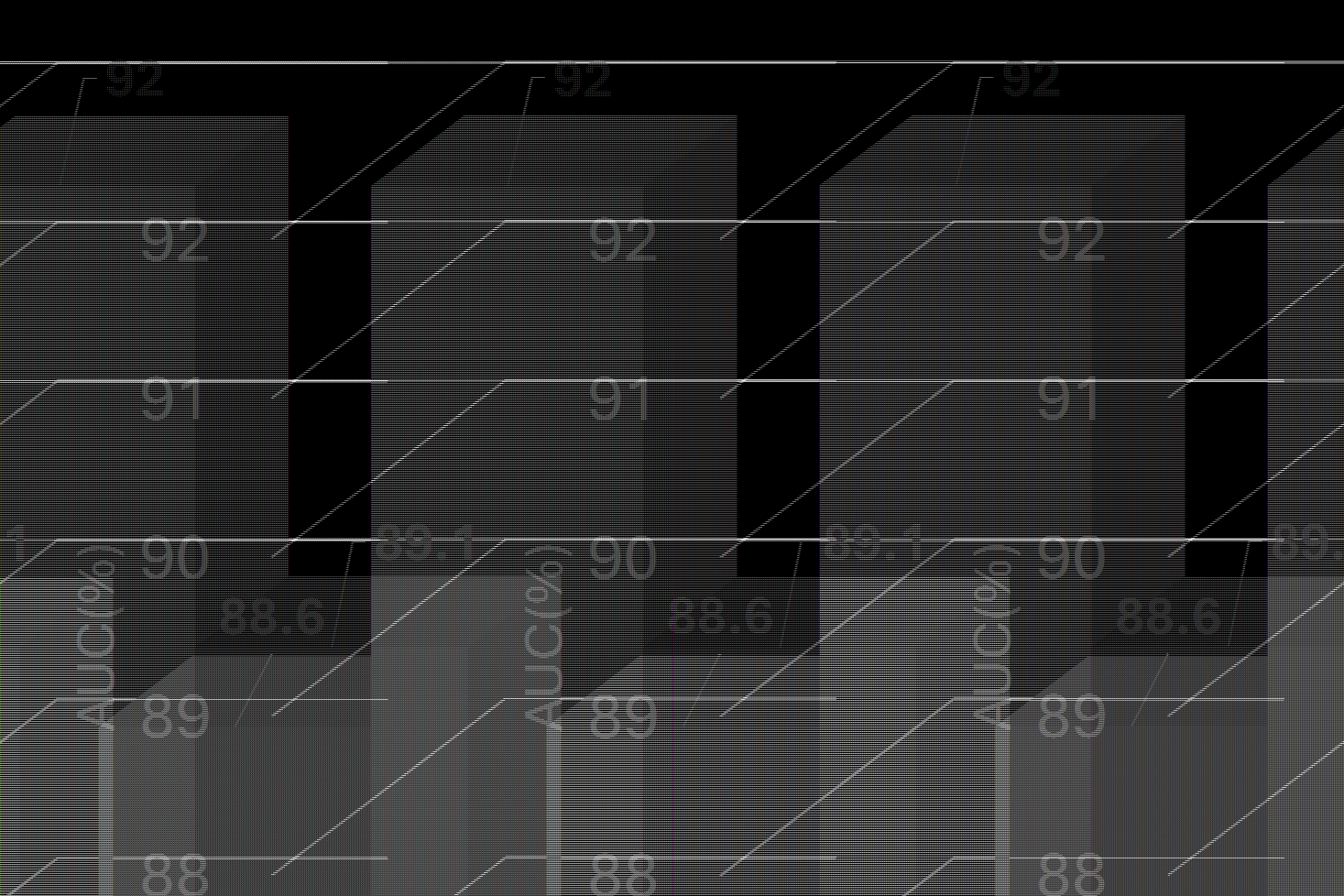}
        \caption{Analysis of the Side Length $p$ of the Sliding Window. }
        \label{fig:hyper0}
    \end{subfigure}
    
    \vspace{4mm} 
    \begin{subfigure}{0.45\columnwidth}
        \centering
        \includegraphics[width=0.9\textwidth]{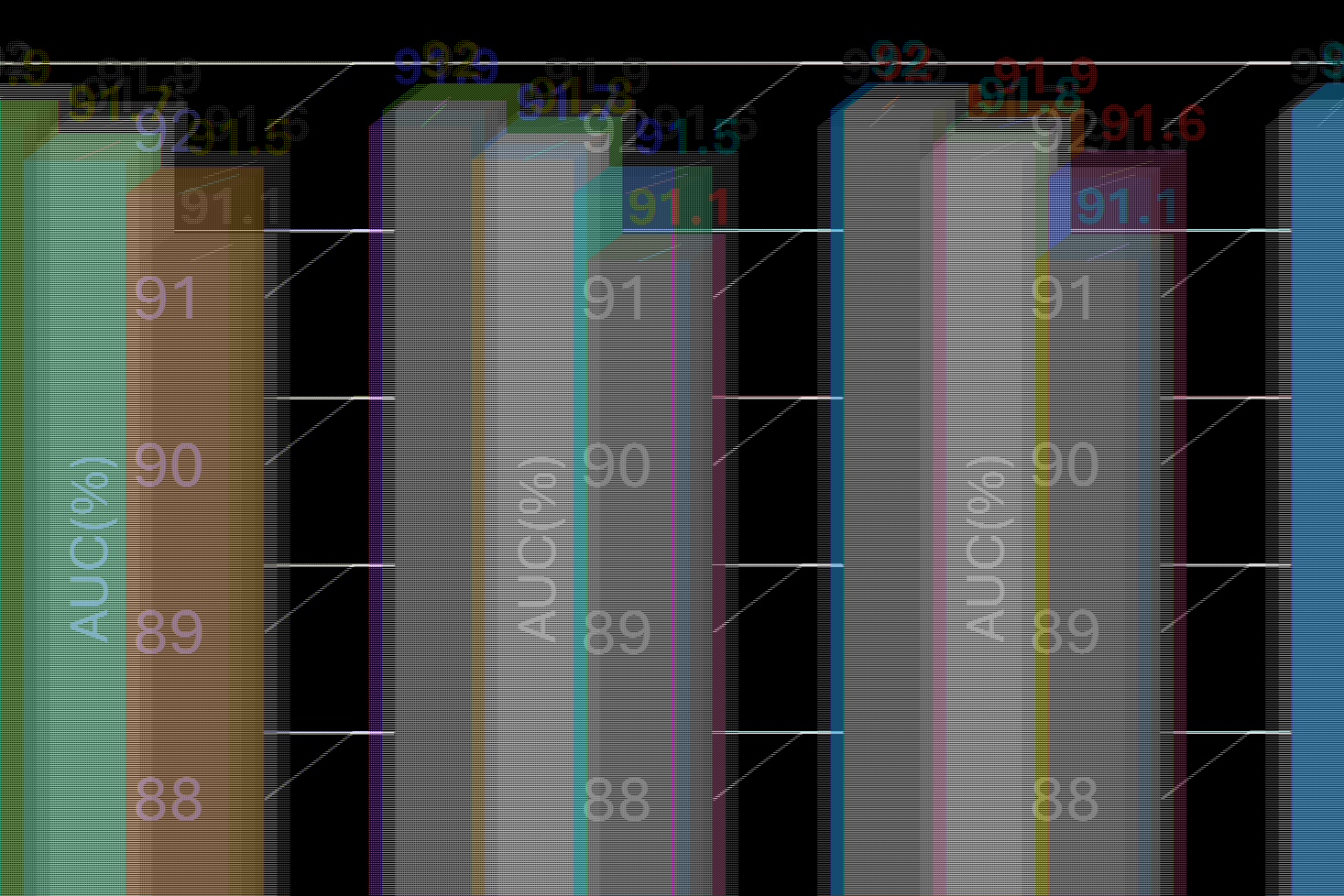}
        \caption{Analysis of $\sigma$ and $k\sigma$}
        \label{fig:hyper2}
    \end{subfigure}
    \hfill
    \begin{subfigure}{0.45\columnwidth}
        \centering
        \includegraphics[width=0.9\textwidth]{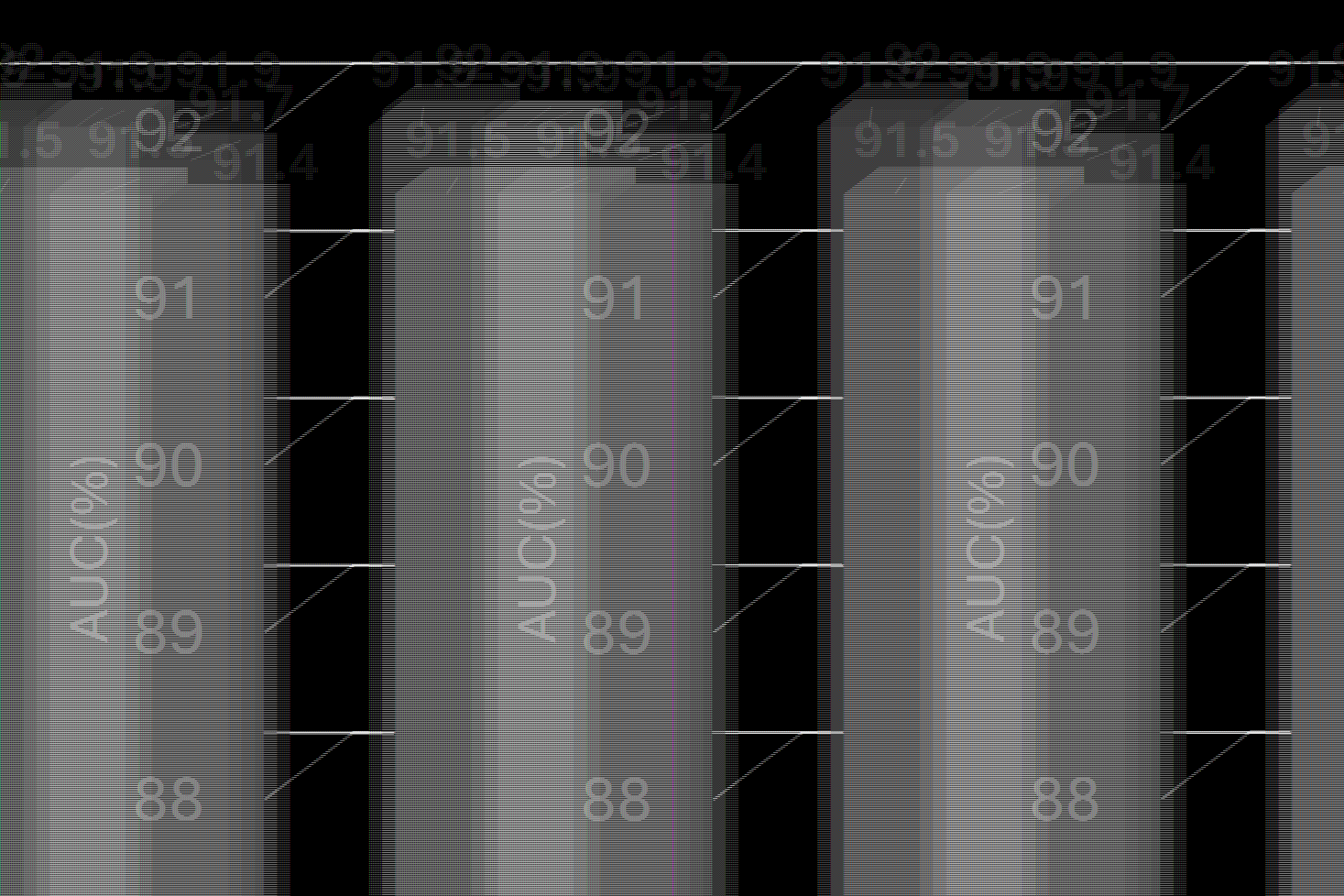}
        \caption{Analysis of $\beta$ and $\gamma$}
        \label{fig:hyper3}
    \end{subfigure}
    \caption{AUC(\%) on ZhangLab Chest X-ray Dataset under Different Hyper-parameters.}
    \label{fig:mainfig}
    \vspace{-5mm} 
\end{figure}

\subsection{Hyper-parameters Analysis}
\label{section:4.6}
In this section, we mainly focus on analyzing the robustness of our model against various hyper-parameters, including the number of prototypes $K$, the weight $\alpha$ of $S_{p-dist}$ for calibrating the anomaly score, the weight $\beta$ of $S_{concen}$ , the weight $\gamma$ of $S_{contra}$, the parameters $\sigma$ and $k\sigma$ of the Gaussian function used in the calculation of $S_{contra}$.

\textbf{Affect of Medical-Prototype Module.}\quad We regard $K$, $\alpha$ and $p$ as hyper-parameters of the Medical-Prototype Module. As shown in Figure~\ref{fig:hyper1}, we find that the number of prototypes has a significant impact on training. Too few prototypes introduce considerable randomness, making it difficult to learn reliable normal features. Conversely, too many prototypes do not enhance performance but increase the model's parameter count and slow down convergence instead. Moreover,as shown in Figure~\ref{fig:hyper0}, the selection of the sliding window's side length $p$ is also crucial. An appropriate $p$ ensures that the model focuses on relevant spatial information while excluding irrelevant interference. At the same time, to fully utilize the ability of $S_{p-dist}$ to calibrate the anomaly score, the weight $\alpha$ is also worth investigating. On the ZhangLab Chest X-ray dataset, we select the number of prototypes $K$ as 10, the weight $\alpha$ of $S_{p-dist}$ as 1 and the sliding window's side length $p$ as 3 in the final model.

\textbf{Effectiveness of Utilizing Contrast Characteristic.}\quad To fully exploit the potential of $S_{contra}$, the parameter $\sigma$ and $k\sigma$ of the Gaussian function should be considered together and carefully selected, as confirmed by Figure~\ref{fig:hyper2} on the ZhangLab Chest X-ray dataset, we select the parameters $\sigma$ and $k\sigma$ of the Gaussian function as 0.6 and 1.2 in the final model.

\textbf{Trade-off between Concentration and Contrast Characteristic.}\quad Both $S_{concen}$ and $S_{contra}$ are strategies for optimizing the anomaly score based on the characteristics of the anomaly map in medical images, and they are used in the same way. Therefore, their weights need to be considered together. In most cases, for general medical datasets, normal images have larger $S_{concen}$ and smaller $S_{contra}$, while abnormal images have the opposite. Thus, the weight $\beta$ for $S_{concen}$ is negative, and the weight $\gamma$ for $S_{contra}$ is positive. For details, refer to Figure~\ref{fig:hyper3}. On the ZhangLab Chest X-ray dataset, we select the weight $\beta$ of $S_{concen}$ as -0.025 and the weight $\gamma$ of $S_{contra}$ as 400 in the final model.

\subsection{Complexity Analysis}
SP-Mamba offers a powerful and efficient approach to medical image anomaly detection tasks by leveraging selective state space models with linear computational complexity and minimal memory usage. Its unique scanning techniques and hardware-aware optimizations make it a promising architecture for handling high-resolution visual data in real-time applications. As shown in Table~\ref{tab:efficiency_1} and Table~\ref{tab:efficiency_2}, our SP-Mamba not only achieves the best anomaly detection performance but also keeps the number of parameters (Params) and floating-point operations (FLOPs) at a relatively low level. Compared to Transformer-based models SQUID~\cite{xiang_squid_2023} and SimSID~\cite{xiang_exploiting_2024}, SP-Mamba requires half or a quarter of the Params and half of the FLOPs, while achieving better performance across various metrics. Specifically, the complexity is only $\frac{1}{50}$ of DiAD~\cite{he2023diaddiffusionbasedframeworkmulticlass}, which is based on a diffusion model. Compared to MambaAD~\cite{he_mambaad_2024}, designed for industrial imaging, our model achieves much better mAD (2.4 improvements) with only 0.1M excessive parameters. This demonstrates the high computational efficiency of our proposed Mamba-based framework, attributed to its hardware-aware optimization and exquisitely designed framework. 


\section{Conclusion}
In this work, we propose SP-Mamba, a spatial-perception Mamba framework for unsupervised medical anomaly detection. In detail, the Circular-Hilbert scanning method and the Medical-Prototype module focus on leveraging spatial position information in anomaly detection tasks, further exploring Mamba's linear modeling capabilities for 2D images and providing valuable insights for future Mamba research. The Anomaly-Scoring module, combining the concentration and contrast characteristics of the medical anomaly map, offers an approach for medical image anomaly maps, with broad applicability in future medical anomaly detection tasks. Extensive experiments on multimodal medical anomaly detection benchmarks demonstrate the effectiveness of our approach in achieving superior performance. Our work emphasizes the importance of spatial information in medical images, and attempts to firstly apply Mamba in medical AD, laying a foundation for future research. SP-Mamba exhibits significant practical implications in enhancing medical imaging, and we hope it can inspire lightweight designs in medical AD in the future. 

\newpage
\begin{acks}
Ruiying Lu would like to thank the Natural Science Basic Research Plan in Shaanxi Province of China under Grant [2024JC-YBQN-0661], Nanning Scientific Research and Technological Development Project (20231042), and Projects of the Industrial Research Plan of Guangxi Institute of Industrial Technology(CYY-HT2023-JSJJ-0021).
\end{acks}

\bibliographystyle{ACM-Reference-Format}
\bibliography{sample-base}


\appendix
\section{More Qualitative Results}

\begin{figure*}[h]
    \centering
    \includegraphics[width=1.75\columnwidth]{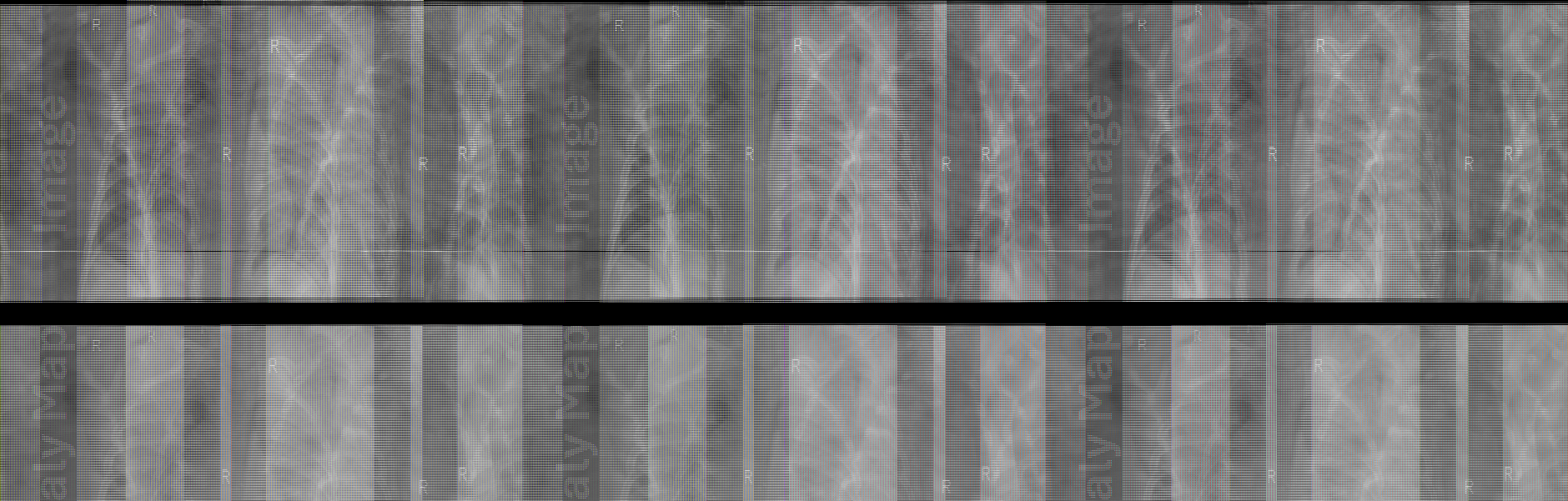}
    \caption{More Qualitative Results on ZhangLab Chest X-ray.}
    \label{zhanglab}
\end{figure*}

\begin{figure*}[h]
    \centering
    \includegraphics[width=1.75\columnwidth]{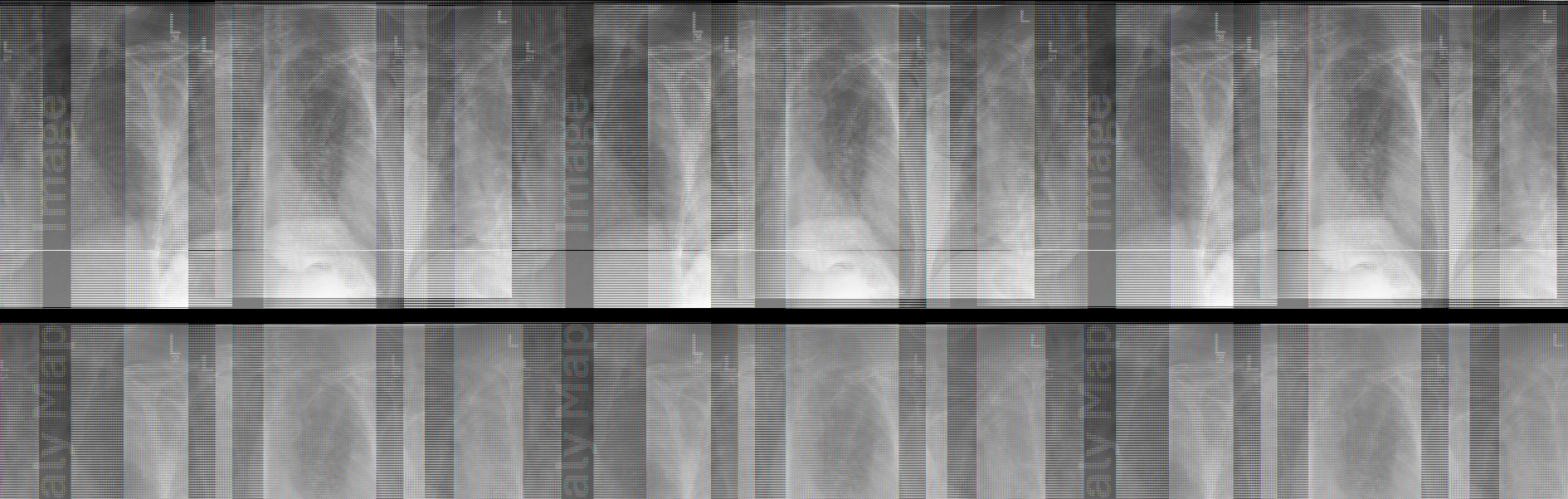}
    \caption{More Qualitative Results on CheXpert.}
    \label{chexpert}
\end{figure*}

\begin{figure*}[h]
    \centering
    \includegraphics[width=1.75\columnwidth]{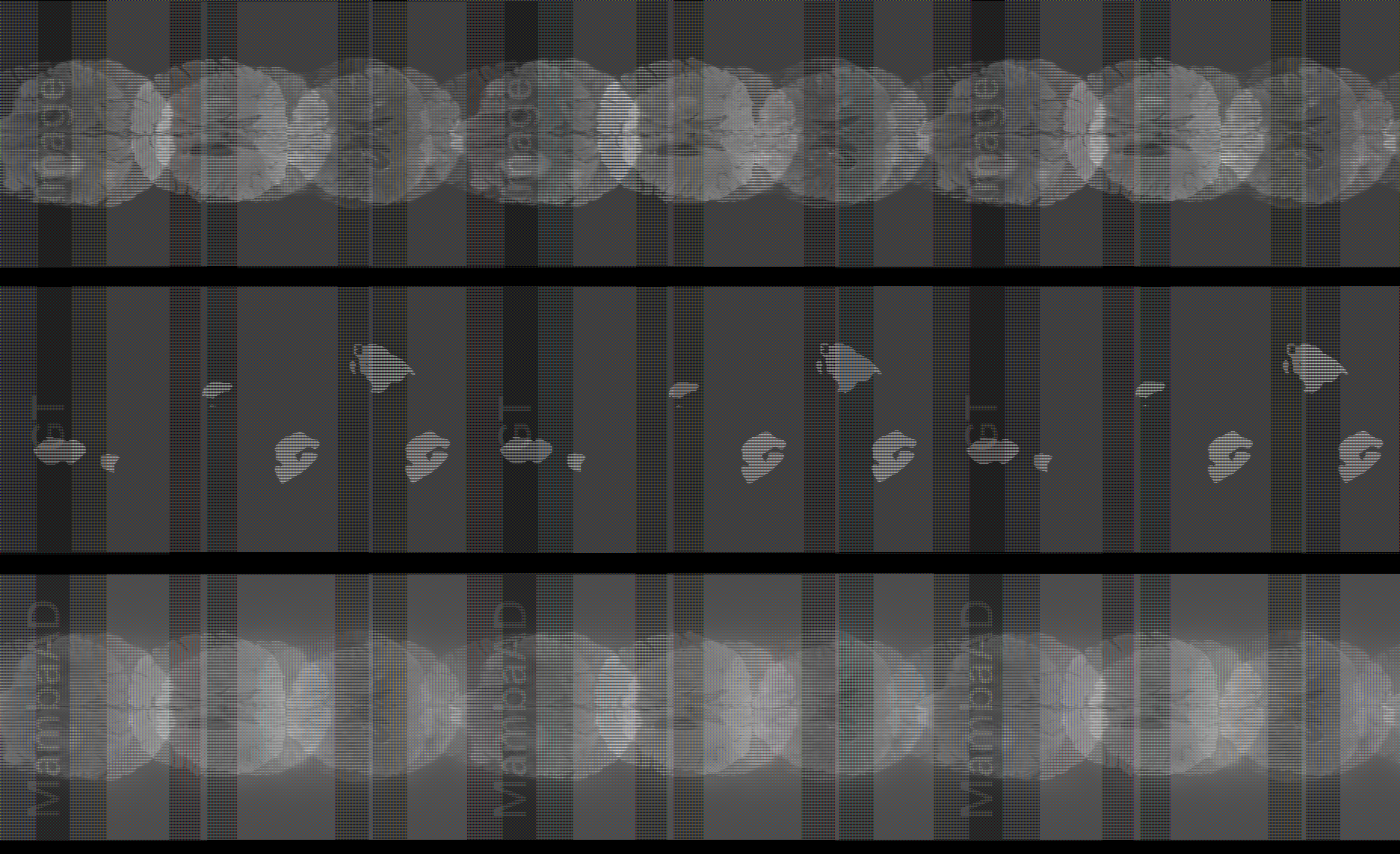}
    \caption{More Qualitative Results on the Brain Class in Uni-Medical.}
    \label{brain}
\end{figure*}

\begin{figure*}[h]
    \centering
    \includegraphics[width=1.75\columnwidth]{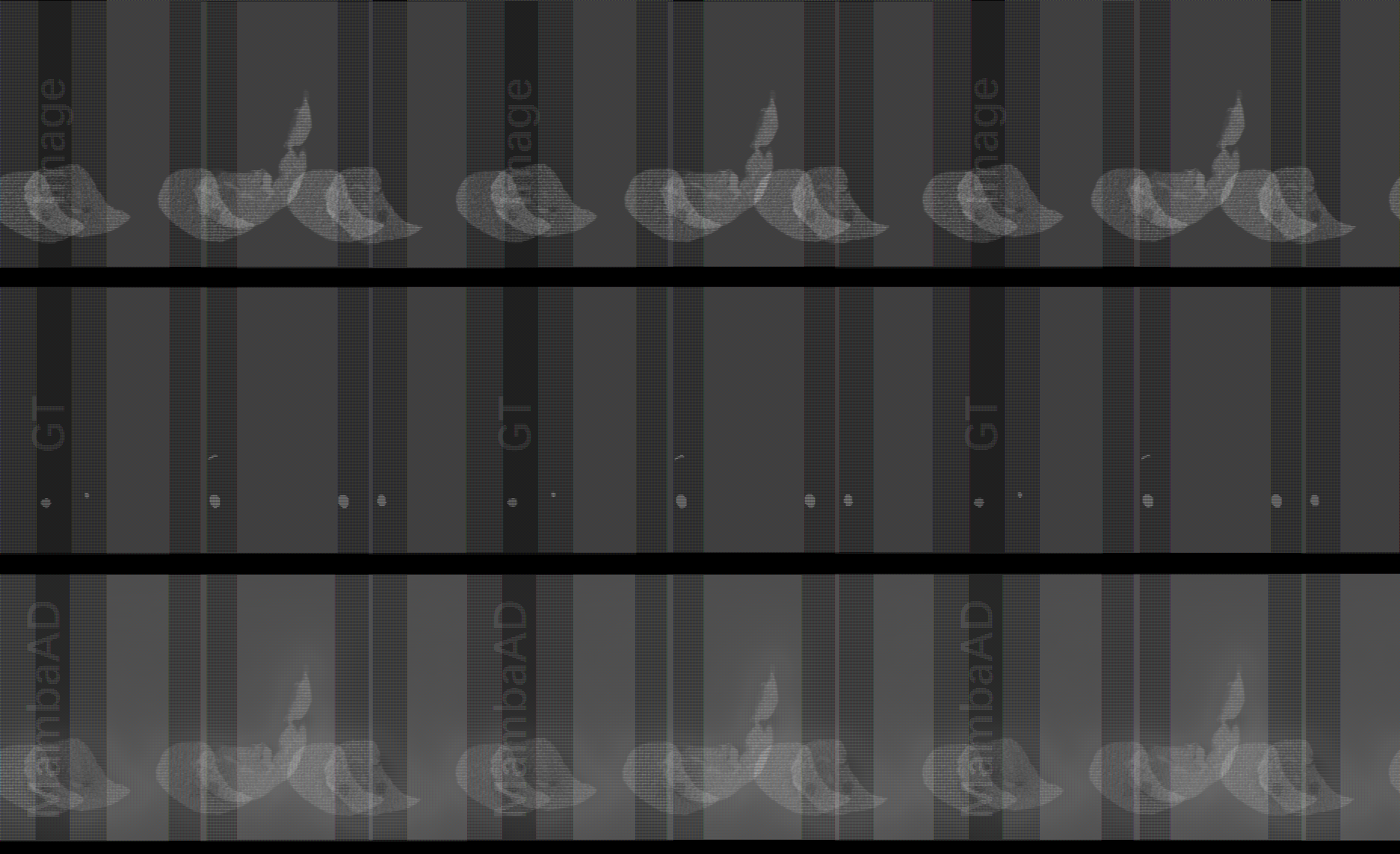}
    \caption{More Qualitative Results on the Liver Class in Uni-Medical.}
    \label{liver}
\end{figure*}

\begin{figure*}[h]
    \centering
    \includegraphics[width=1.75\columnwidth]{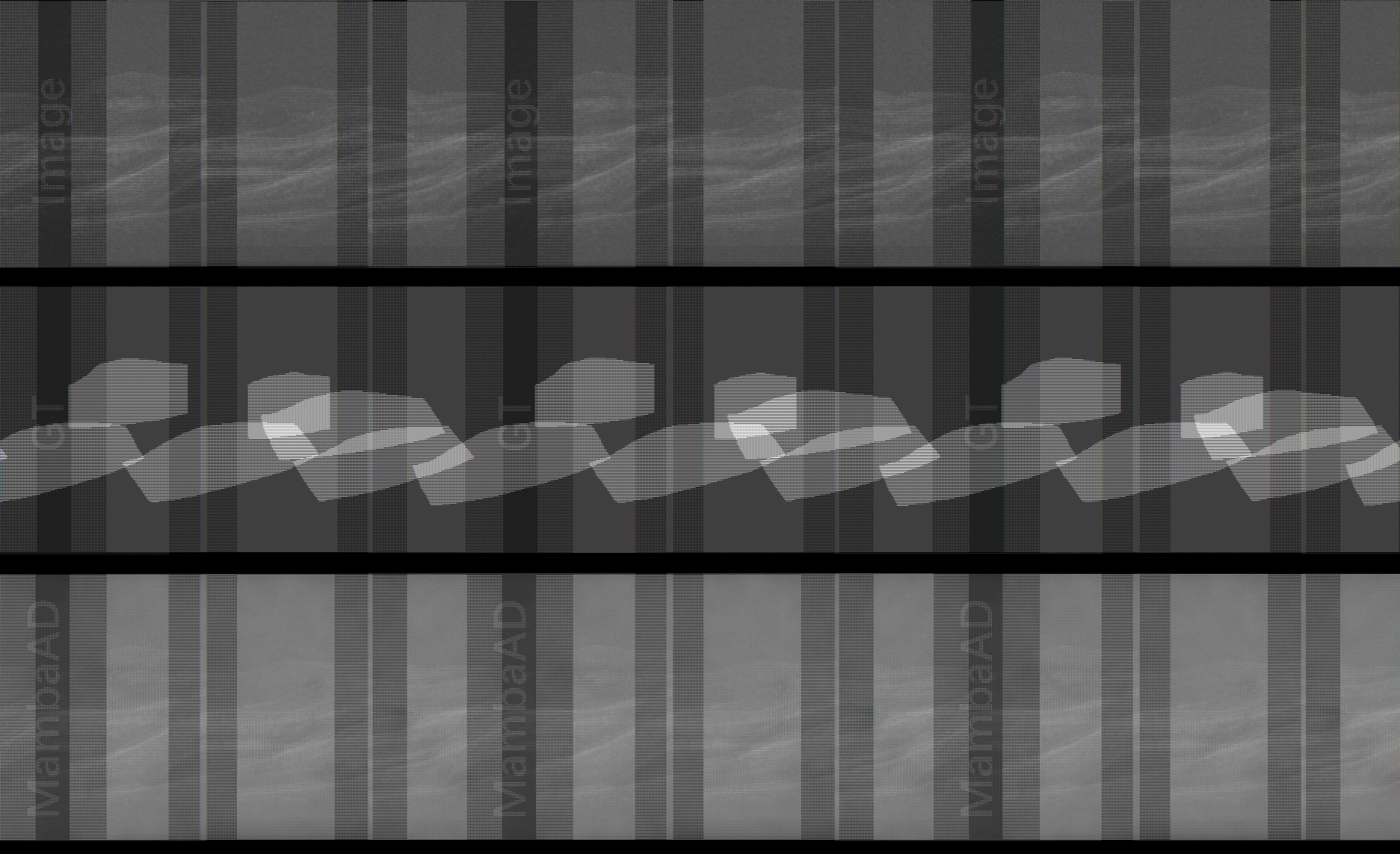}
    \caption{More Qualitative Results on the Retinal Class in Uni-Medical.}
    \label{retinal}
\end{figure*}

\end{document}